\documentclass{article}

\usepackage{microtype}
\usepackage{graphicx}
\usepackage{subfigure}
\usepackage{booktabs} 
\usepackage{array}
\usepackage{multirow}

\usepackage{makecell}

\usepackage{hyperref}

\usepackage[preprint]{icml2025}

\usepackage{amsmath}
\usepackage{amssymb}
\usepackage{mathtools}
\usepackage{amsthm}

\usepackage[capitalize,noabbrev]{cleveref}

\theoremstyle{plain}

\theoremstyle{definition}

\newtheorem*{definition*}{Definition}

\theoremstyle{remark}

\usepackage[textsize=tiny]{todonotes}

\icmltitlerunning{Scaling Flaws of Verifier-Guided Search}

\begin{document}

\twocolumn[
\icmltitle{
Scaling Flaws of Verifier-Guided Search in Mathematical Reasoning
}



\icmlsetsymbol{equal}{*}

\begin{icmlauthorlist}
\icmlauthor{Fei Yu}{yyy}
\icmlauthor{Yingru Li}{yyy}
\icmlauthor{Benyou Wang}{yyy}
\end{icmlauthorlist}

\icmlaffiliation{yyy}{The Chinese University of Hong Kong, Shenzhen, China}

\icmlcorrespondingauthor{Benyou Wang}{wangbenyou@cuhk.edu.cn}
\icmlcorrespondingauthor{Yingru Li}{szrlee@gmail.com}

\icmlkeywords{Machine Learning, ICML}

\vskip 0.3in
]



\printAffiliationsAndNotice{}  

\begin{abstract}
Large language models (LLMs) struggle with multi-step reasoning, where inference-time scaling has emerged as a promising strategy for performance improvement. Verifier-guided search outperforms repeated sampling when sample size is limited by selecting and prioritizing valid reasoning paths. However, we identify a critical limitation: scaling flaws, prevalent across different models (Mistral 7B and DeepSeekMath 7B), benchmarks (GSM8K and MATH), and verifiers (outcome value models and process reward models). As sample size increases, verifier-guided search exhibits diminishing advantages and eventually underperforms repeated sampling. Our analysis attributes this to verifier failures, where imperfect verifiers misrank candidates and erroneously prune all valid paths. These issues are further exacerbated in challenging and out-of-distribution problems, restricting search effectiveness. To mitigate verifier failures, we explore reducing reliance on verifiers and conduct preliminary investigations using two simple methods. Our findings reveal fundamental limitations in verifier-guided search and suggest future directions.
\end{abstract}

\section{Introduction}
Multi-step reasoning is challenging to LLMs~\cite{MATH21, MINIF2F22}. 
Recent studies have identified inference-time scaling~\cite{RS24, Snell24, wu2024inference} as a promising strategy to enhance LLM performance on multi-step reasoning. By increasing inference-time computation through multiple attempts via repeated sampling~\cite{RS24}, LLMs can solve more problems, with at least one attempt succeeds. Building on this insight, search-based approaches have emerged to guide computation toward more effective reasoning paths~\cite{Snell24, wu2024inference}. 

Search reallocates computational resources by evaluating and selecting partial paths during generation. A common approach for path evaluation uses verifiers~\cite{Snell24, wu2024inference}, such as outcome value models (OVMs)~\cite{OVM23} and process reward models (PRMs)~\cite{PRM24}, to score and rank candidates, prioritizing valid paths. This makes verifier-guided search effective for challenging problems with sparse valid solutions, offering advantages over repeated sampling when the sample size is limited.

\textbf{Obvervation of scaling flaws}. 
However, we observe that verifier-guided search (e.g. OVM- and PRM-guided) might 
experience diminishing advantages and eventually underperforms repeated sampling as the sample size scales. Its performance improves more slowly than repeated sampling, ultimately making them less effective. We refer to this phenomenon as \emph{scaling flaws of verifier-guided search}. 

\textbf{Identification of verifier failures}. 
To understand the cause of scaling flaws, we analyze search failures and identify verifier selection failures as the main factor, where imperfect verifiers misrank and incorrectly prune all valid paths—an issue we term ``verifier failures''. Morever, verifier selection itself exhibits scaling issues: as the candidate size increases, valid paths become more widespread across the problem set, yet verifiers struggle to identify them, leading to their erroneous pruning. This contributes to the overall search scaling flaws.

\textbf{Analysis of verifier failures}. 
Our investigation shows that verifier failures and scaling flaws worsen in challenging and out-of-distribution problems. As problem difficulty and solution sparsity increase, scaling flaws intensify. This paradoxically undermines search, which is intended to outperform repeated sampling in such cases. Moreover, out-of-distribution problems, common in real-world deployment, exacerbate these challenges, highlighting fundamental limitations of verifier-guided search approaches.

\textbf{Mitigating verifier failures}. 
To explore potential approaches for mitigating verifier failures, we conduct a preliminary investigation into two simple methods that reduce reliance on verifiers, both of which demonstrate benefits.

\textbf{Summary of contributions}. (1) This work identifies and analyzes the scaling flaws of verifier-guided search (2) We pinpoint verifier failures as the primary cause of these flaws (3) Our analysis reveals that these issues become more severe for challenging and out-of-distribution problems, raising concerns about the development of verifier-guided search algorithms and their application in real-world settings (4) We suggest reducing reliance on verifiers and conduct preliminary investigations using two simple methods.

\section{Related Works}

\paragraph{Search algorithms} 
Search algorithms often face a tradeoff between effectiveness and efficiency. Approaches like Monte Carlo Tree Search~\cite{RAP23,tian24} improve effectiveness by incorporating backtracking, but at the cost of efficiency. Other methods prioritize efficiency with minimal sacrifice in effectiveness~\cite{rebase24}. In this work, we use a simple beam search algorithm~\cite{OVM23,AlphaMath24} for our experiments, focusing on highlighting challenges in the candidate evaluation and selection stage, orthogonal to these advanced techniques.


\paragraph{Candidate evaluation in search} 
Candidate evaluation is a crucial stage that determines which paths are more valuable for further selection and exploration. Some methods rely on the some rule-based heuristics~\cite{deepseekprover15-24}, with limited effectiveness. Some approaches involve lookahead techniques to assess candidates by simulating their subsequent outcomes~\cite{Snell24,Wan24}, which significantly increases computational cost. Other methods incorporate external verifier models~\cite{OVM23,Snell24} to evaluate each candidate. In this work, we focus on the challenges and limitations of the this approach.


\section{Background: Verifier-Guided Search}
\label{sec:scaling_flaws}

\begin{figure*}[!ht]
    \centering
    \subfigbottomskip=2pt
    \subfigcapskip=-2pt
    \subfigure[\label{fig:gsm8k_scaling_flaws_beam_ovm}OVM on GSM8K]{
        \includegraphics[width=0.4\linewidth]{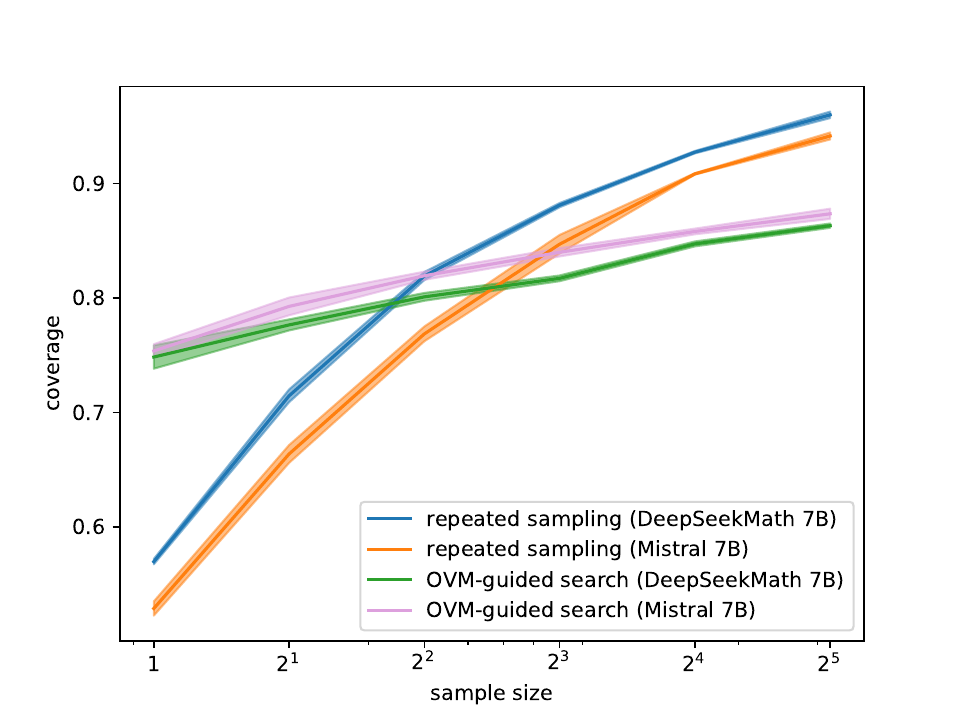}}
    \subfigure[\label{fig:gsm8k_scaling_flaws_beam_prm}PRM on GSM8K]{
        \includegraphics[width=0.4\linewidth]{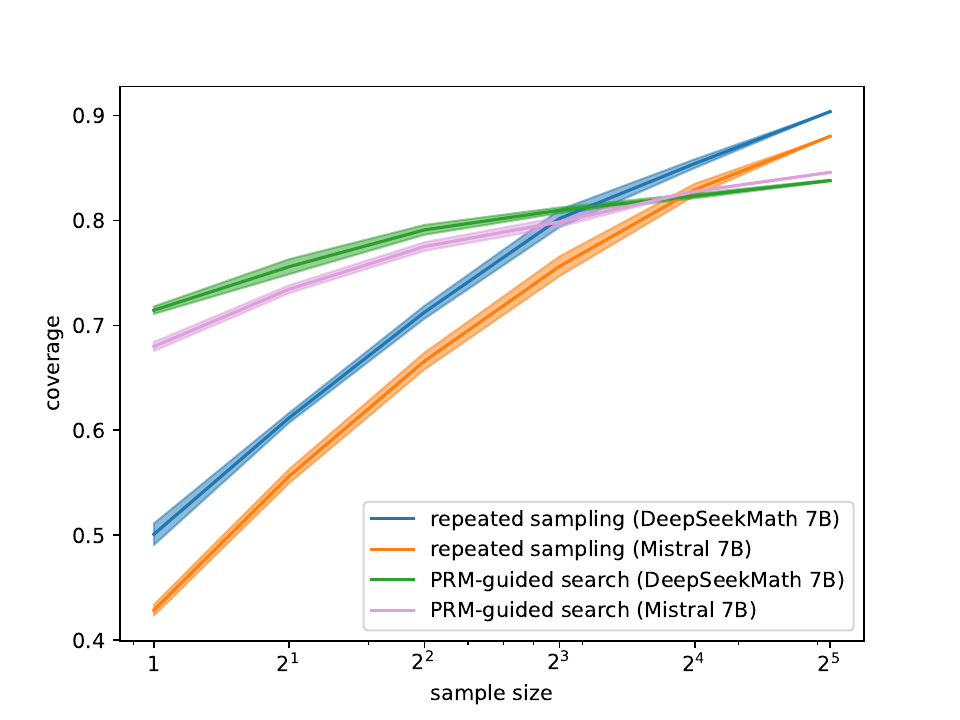}}
        
    \subfigure[\label{fig:math_scaling_flaws_beam_ovm}OVM on MATH]{
        \includegraphics[width=0.4\linewidth]{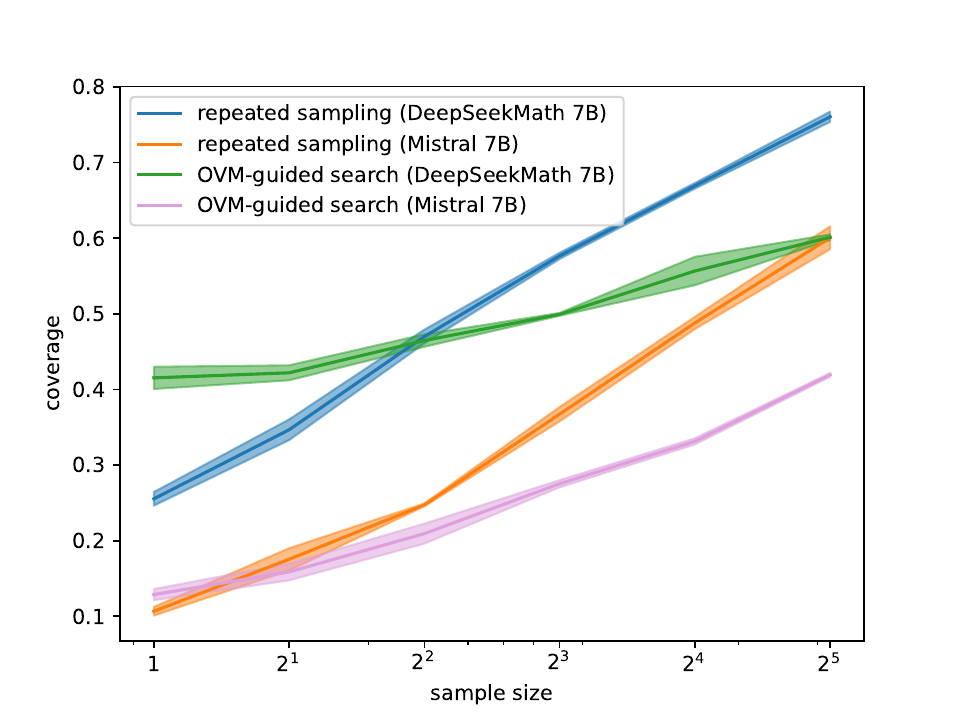}}
    \subfigure[\label{fig:math_scaling_flaws_beam_prm}PRM on MATH]{
        \includegraphics[width=0.4\linewidth]{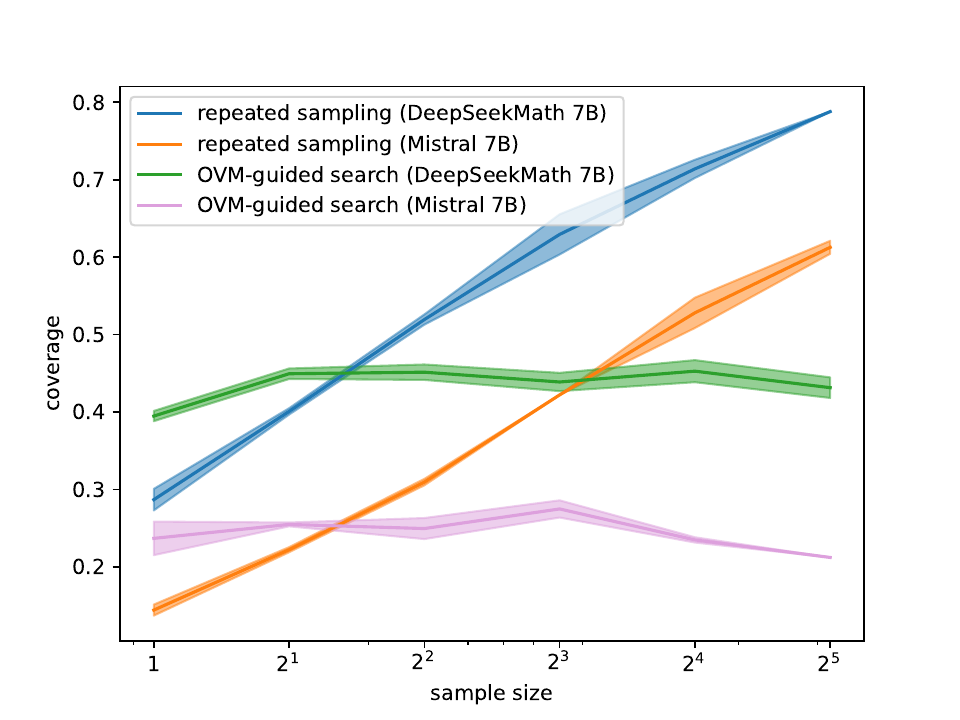}}
    \caption{\label{fig:scaling_flaws_beam}Scaling Flaws in OVM-guided search and PRM-guided search on GSM8K and MATH (scaling of sample sizes). While verifier-guided search outperforms repeated sampling initially, its performance increases at a slower rate, ultimately underperforming repeated sampling.}
\end{figure*}

This section begins by defining mathematical reasoning questions and introducing two widely employed solution frameworks: repeated sampling and search. We then detail a specific search framework, beam search, and discuss two widely-used verifier types employed in the search process.

\begin{definition*}
    \textbf{A mathematical reasoning question} \textit{$q$ requires a step-by-step solution path $S=[s^1, \dots, s^T, a]$ to be addressed, where $s^i$ represents the i-th step, $T$ is the number of steps, and $a$ is the final answer.} 
\end{definition*}

Multi-step reasoning~\cite{GSM8K21,MATH21} suffers from error propagation issues--errors in earlier steps affect later ones, compromising the final answer. Recent studies show that LLMs can address more challenging problems through repeated sampling~\cite{RS24}.

\paragraph{Repeated sampling} 
LLMs can solve some challenging problems through multiple attempts~\cite{GSM8K21,RS24}, i.e. repeatedly sampling a set of solution paths $\bigl\{S_k \bigl\}_{k=1}^{K}$ from the generator. Increasing the number of attempts, $K$, often improves the coverage—the fraction of problems for which at least one sampled path is correct, but also requires more computation.

However, repeated sampling becomes inefficient for challenging problems, like competition-level mathematics problems~\cite{MATH21}, where it often demands many more attempts to find a correct solution~\cite{RS24}.


\subsection{Search} 

Search aims to explore correct solutions more efficiently than repeated sampling by pruning unpromising partial paths and discarding early errors. This paper focuses on \emph{step-level beam search}, a widely used and sufficiently straightforward framework for illustrating the core concept.


\paragraph{Step-level beam search} 
This framework intervenes in generation and selection at the step level and explore multiple paths in parallel. Given a question $q$, at each step $t$, the generator produces $K$ candidates $\mathbb{S}^{(1:t)} = \bigl\{S^{(1:t)}_k \bigl\}_{k=1}^{K}$, where $S^{(1:t)}_k=[s^1_k,\dots,s^t_k]$ is the $k$-th partial path. During the selection stage, a scoring function $f$ evaluates these candidates, assigning scores $\mathbb{V}^{(1:t)} = \bigl\{v^{(1:t)}_k \bigl\}_{k=1}^{K}$, where $v^{(1:t)}_k$ is the score for $S^{(1:t)}_k$, ranking them for selection. The top $b$ paths proceed to the next step, generating $K/b$ new candidates each, maintaining a total of $K$ candidates. This process repeats until all $b$ paths terminate, yielding $b$ full solution paths. See details in~\cref{algo:beam_search}. The hyperparameter $b$ controls the number of parallel paths. Larger $b$ or $K$ improve the ability to handle a wider range of problems.

Search using verifiers as scoring functions is particularly noteworthy~\cite{OVM23,AlphaMath24,Snell24}. We refer to this approach as ``verifier-guided search''.






\subsection{Verifiers} 

Verifiers~\cite{PRM24,OVM23} are commonly employed as scoring functions to evaluate candidate, determining which paths to be further explored. In this work, we focus on the two most widely used types of verifiers--Outcome-supervised Value Models~\cite{OVM23} and Process-supervised Reward Models~\cite{PRM24}.

\paragraph{Outcome-Supervised Value Model (OVM)}
The OVM~\cite{OVM23} evaluates each candidate by estimating the probability of arriving at a correct answer from the given partial path. It assumes that each local step with the highest probability of success ultimately leads to the correct answer. We refer to search using OVM for evaluation as ``OVM-guided search''.

\paragraph{Process-Supervised Reward Model (PRM)}
The PRM~\cite{PRM24} evaluates each candidate by predicting its step correctness. It assumes that each correct local step guides to the correct final answer. We refer to search using PRM for evaluation as ``PRM-guided search''.

Verifiers play a key role in candidate evaluation and selection, directly influencing the search success. When they correctly identify valid paths, they can steer the search towards correct solutions more efficiently than repeated sampling. 

However, we observe that although the search process initially shows advantages over repeated sampling, these advantages disappear as scaling, as shown in the next section.


\section{Scaling Flaws of Verifier-Guided Search}
\label{sec:flaws}

In this section, we present extensive experiments showing that verifier-guided search suffers \emph{scaling flaws}: it outperforms repeated sampling at small sample sizes but underperforms it at large sample sizes. These flaws are worse on more difficult and out-of-distribution problems.

\subsection{Experimental Setup}

\paragraph{Benchmarks} 
We perform experiments on two mathematical reasoning datasets: GSM8K~\cite{GSM8K21} and MATH~\cite{MATH21}. The experiments are conducted under four distinct settings, including two in-distribution and two out-of-distribution (OOD) scenarios, as detailed below:
\begin{itemize}
    \item \textbf{GSM8K}: The official training split is used for training, and the model is evaluated on the test split.
    \item \textbf{MATH}: The official training split, comprising 7,500 problems, is used for training, while evaluation is performed on the MATH500~\cite{PRM24}.
    \item \textbf{OOD-L4}: Training is conducted on MATH Level 1, Level 2, Level 3, and Level 5 problems, while evaluation is performed specifically on Level 4 problems within MATH500. This setting requires models to generalize to problems of median difficulty.
    \item \textbf{OOD-L5}: We train on MATH Level 1 - Level 4 problems and evaluate on Level 5 problems within MATH500. In this setting, models are required to generalize to solve more difficult problems.
\end{itemize}

\paragraph{Models} 
We use Mistral 7B~\cite{Mistral7B-23} and DeepSeekMath 7B~\cite{DeepSeekMath24} for the GSM8K and MATH experiments, and exclusively use DeepSeekMath 7B for the two OOD settings. For each setting, the base models are trained on the corresponding training sets to serve as the generators. The OVMs used in each setting are initialized from these generators. For PRMs, we leverage the open-source Math-Shepherd dataset~\cite{Math-Shepherd24}. Generators are first fine-tuned on a subset of this data, after which PRMs, initialized from the corresponding generators, are trained under supervision using process labels.

\paragraph{Scaling beam search} 
We investigate the scaling laws of two factors: (1) the number of parallel explored paths $b$, with $K/b$ fixed at 8, and (2) the number of generated candidates $K$, with $b$ fixed at 8. For the comparison between beam search and repeated sampling, we align them in terms of ``sample size'', which represents the number of complete solution paths generated by each algorithm. For beam search, the sample size corresponds to the number of parallel explored paths, $b$, while for repeated sampling, it corresponds to the number of attempts. Each experiment is repeated three times, and we report the average coverage (i.e. the fraction of problems for which at least one sampled path is correct) along with their standard deviation.

See implementation details in~\cref{app:implementation}.

\subsection{Scaling Flaws}

\begin{figure*}[!ht]
    \centering
    \subfigbottomskip=2pt
    \subfigcapskip=-2pt
    \subfigure[\label{fig:math_scaling_flaws_candidate_ovm}OVM on MATH]{
        \includegraphics[width=0.4\linewidth]{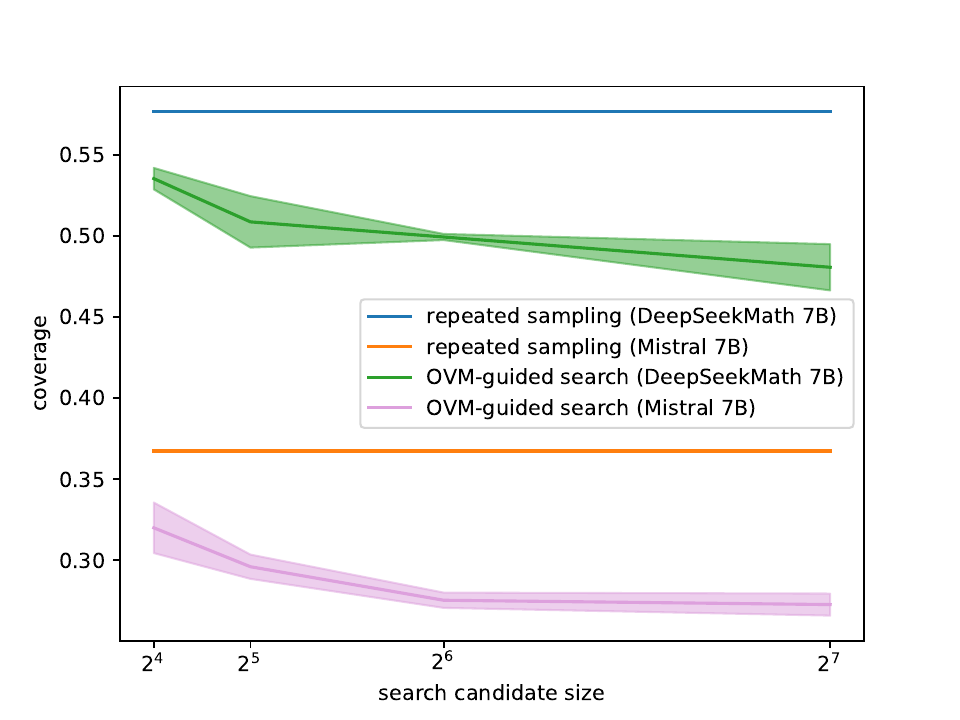}}
    \subfigure[\label{fig:math_scaling_flaws_candidate_prm}PRM on MATH]{
        \includegraphics[width=0.4\linewidth]{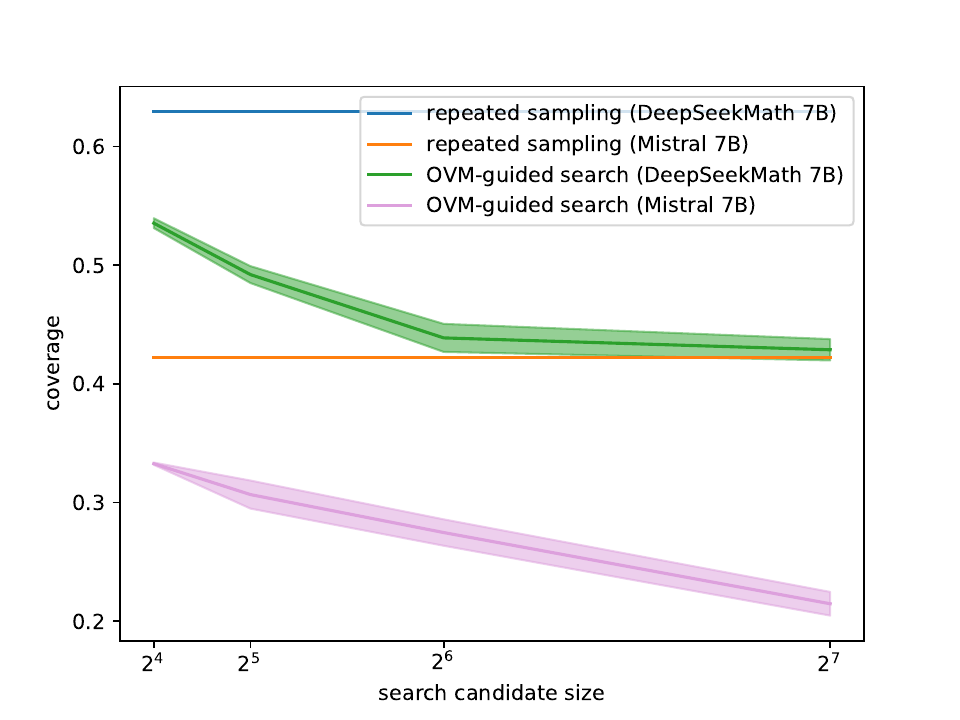}}

    \subfigure[OVM on OOD-L5]{
        \includegraphics[width=0.4\linewidth]{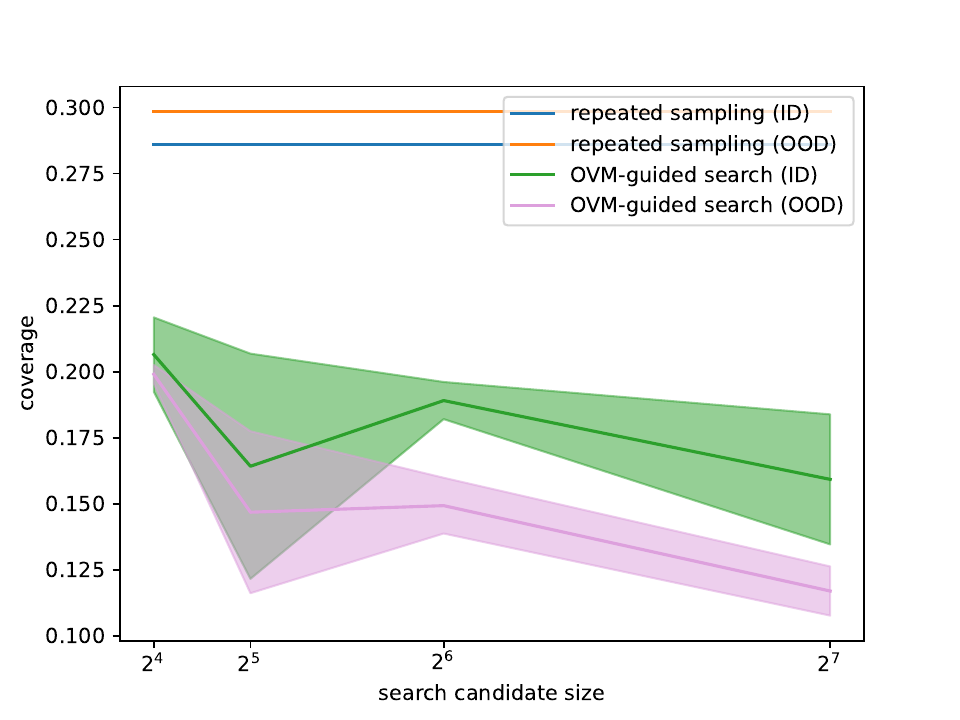}}
    \subfigure[PRM on OOD-L5]{
        \includegraphics[width=0.4\linewidth]{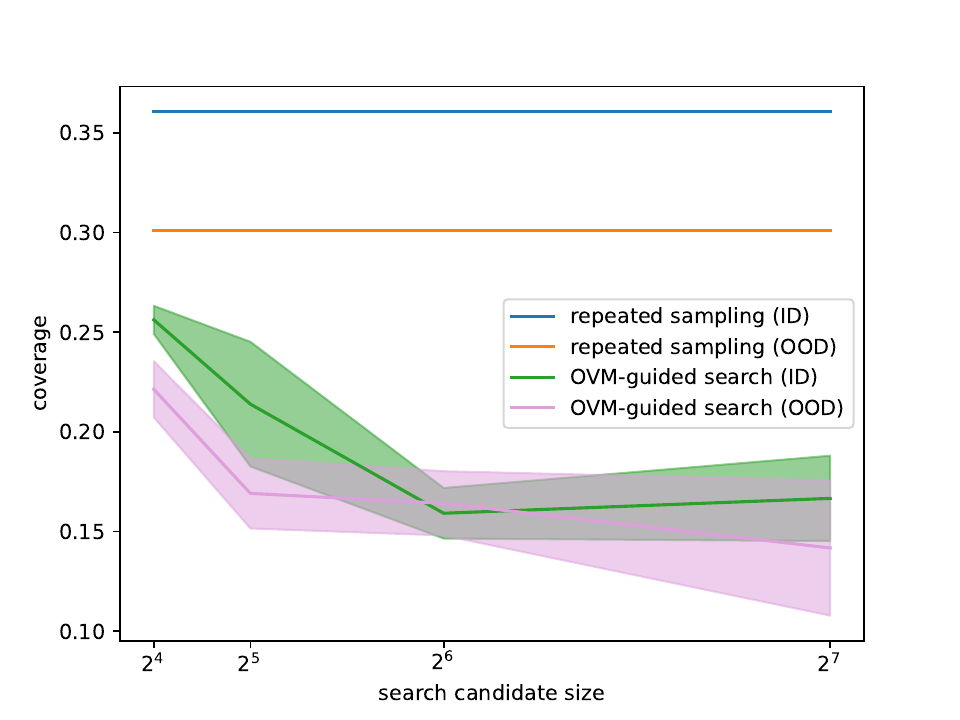}}
    \caption{\label{fig:scaling_flaws_candidate}Scaling Flaws in OVM-guided search and PRM-guided search on MATH and OOD-L5 (scaling generated candidate size).}
\end{figure*}

The results of scaling verifier-guided search are presented in~\cref{fig:scaling_flaws_beam}-~\cref{fig:scaling_flaws_candidate}. Notably, both OVM-guided and PRM-guided encounter scaling flaws across all settings. 

\paragraph{Scaling flaws of verifier-guided search} 
\textit{Verifier-guided search encounters scaling flaws across benchmarks and models}. Both OVM-guided and PRM-guided search exhibit failures in scaling sample sizes (\cref{fig:scaling_flaws_beam}) and generated candidate sizes (\cref{fig:scaling_flaws_candidate}). When scaling sample sizes, as shown in~\cref{fig:scaling_flaws_beam}, both OVM-guided and PRM-guided search initially outperforms repeated sampling, e.g. when the sample size is set to 1, on GSM8K and MATH. However, as the sample size increases, the performance of verifier-guided search increases at a slower rate compared to repeated sampling, ultimately underperforming repeated sampling. 

For instance, in~\cref{fig:gsm8k_scaling_flaws_beam_prm}, PRM-guided search based on either DeepSeekMath or Mistral initially achieves over 20\% higher performance than repeated sampling when the sample size is 1. However, this advantage erodes as the sample size scales, and by a sample size of 16, verifier-guided search becomes inferior to repeated sampling, reaching approximately 5\% lower performance when scaled to 32. Similarly, in~\cref{fig:math_scaling_flaws_beam_ovm}, OVM-guided search based on DeepSeekMath or Mistral is overtaken by repeated sampling at a sample size of 4, eventually falling behind by approximately 20\% when the sample size reaches 32. 

Moreover, increasing the number of generated candidates fails to improve and even degrades the performance of verifier-guided search, as shown in~\cref{fig:scaling_flaws_candidate}.

\paragraph{Intensified on difficult problems} 
\textit{Scaling flaws are more severe on more difficult problems}. As shown in~\cref{fig:scaling_flaws_beam} and~\cref{tab:scaling_flaws_diff_beam}, scaling flaws are more pronounced on MATH than on GSM8K and become increasingly severe as problem difficulty increases within MATH. In~\cref{fig:scaling_flaws_beam}, the performance degradation—measured as the gap between search and repeated sampling at a sample size of 32—is approximately 10\% for OVM-guided search (both DeepSeekMath and Mistral) on GSM8K, increasing to around 20\% on MATH. Similarly, for PRM-guided search, the performance degradation rises from about 5\% on GSM8K to nearly 30\% on MATH. 

Furthermore, as observed in~\cref{tab:scaling_flaws_diff_beam}, consistent with previous research~\cite{Snell24}, verifier-guided search shows greater benefits over repeated sampling for moderate problems. For instance, at a sample size of 1, gains are larger for Level 2–Level 4 problems compared to Level 1 and Level 5. However, as problem difficulty increases, the performance degradation of both OVM- and PRM-guided search approximately upward monotonically. Notably, the loss gap exceeds 20\% when comparing Level 1 to Level 5 problems, suggesting the increasing severity of scaling flaws.

\begin{table}[h]
\caption{\label{tab:scaling_flaws_diff_beam}Increased average coverage of search over repeated sampling across various problem difficulties on MATH and OOD settings (DeepSeekMath 7B). `L': `Level', \#sample: sample size.}
\vskip 0.1in
\scriptsize
\begin{center}
\setlength{\tabcolsep}{0.8mm}
\begin{tabular}{lrrrrrrrrr}
\toprule
                     & \#sample      & L1     & L2     & L3      & L4      & L5    & OOD-L4 & OOD-L5 \\
\midrule
\multirow{2}{*}{OVM} & 1   & 11.6\%  & 23.0\%  & 21.3\%   & 17.7\%  & 7.0\% & 14.1\% & 1.2\% \\
                     & 32  & -3.1\%  & -6.0\%  & -15.6\%  & -19.3\% & -23.9\% & -25.8\% & -32.8\% \\
\multirow{2}{*}{PRM} & 1   & 8.5\%   & 18.9\%  & 12.1\%   & 13.5\%  & 2.5\% & 8.1\% & 5.0\% \\
                     & 32  & -11.6\% & -24.4\% & -42.5\%  & -43.8\% & -37.8\% & -46.1\% & -39.1\% \\
\bottomrule
\end{tabular}
\end{center}
\vskip -0.1in
\end{table}

\paragraph{Intensified on OOD problems} 
\textit{Scaling flaws are more severe on OOD problems}. As shown in~\cref{tab:scaling_flaws_diff_beam}, performance degradation at a sample size of 32 is more pronounced in OOD settings for both OVM- and PRM-guided search. For instance, the performance degradation of OVM-guided search on the in-distribution Level 4 setting is 19.3\%, and it increases to 25.8\% in the OOD-L4 setting. Similarly, the loss rises from 23.9\% on Level 5 to 32.8\% in the OOD-L5 setting. These results reveal the exacerbated impact of scaling flaws when generalizing to OOD problems.

A notable concern arises: these findings indicate that the performance degradation of verifier-guided search compared to repeated sampling as scaling is enhanced with increasing problem difficulty. However, this contradicts the purpose of verifier-guided search, which is designed to improve performance in solving difficult problems. Furthermore, out-of-distribution scenarios—commonly encountered in real-world deployment—further exacerbate these scaling flaws.

\section{Verifier Failures}
\label{sec:verifier_flaws}

Section~\ref{sec:flaws} observed scaling flaws in verifier-guided search, but the underlying cause remains unknown. This section conducts an in-depth analysis, identifying incorrect selection due to imperfect verifiers as the root cause of these flaws. 
In~\cref{sec:verifier_flaws_scaling}, we term this phenomenon as ``verifier failures'' and analyze its connection to search scaling flaws. In~\cref{sec:challenging_block_more}, we investigate the distribution of failed selection stages during the search, examining their correlation with the sparsity of candidate space.


\subsection{Experimental setup} 
In this section, we analyze the selection stages from two perspectives: (1) only the first selection stage with a large number of candidates $K=256$ to study the relationship between the number of candidates and the performance of verifier selection, including both OVM selection and PRM selection (2) analyze all selection stages during the OVM-guided search with $b=8, K=64$, as this configuration suffers from scaling flaws across benchmarks and models while maintaining an acceptable computational cost for valid path labeling.

The selection stages during the search are analyzed based on a single criterion: whether at least one valid path is selected when valid paths are available. A candidate is considered a valid path if it can lead to the correct final answer. To determine valid paths, we complete each partial path by rolling out multiple samples and verifying whether any of the rollouts successfully reach the correct answer. Specifically, we generate 4 rollouts per candidate for GSM8K and 16 rollouts for MATH and OOD settings.


\subsection{Verifier Failures Cause Search Scaling Flaws}
\label{sec:verifier_flaws_scaling}
Search failures can arise from either the generation stage or the selection stage—specifically, when no valid candidates are generated or when valid paths produced during generation fail to be selected.

\paragraph{Generation vs.\ selection failures}
\textit{Search failures are largely attributable to selection failures}. We analyze all search processes in which problems fail to be solved and attribute these failures to either generation or selection. A failure is attributed to the generation stage if there is at least one intermediate step where no valid partial paths are generated. Conversely, it is attributed to the selection stage if, at any intermediate step, valid paths are produced but fail to be selected. As shown in~\cref{tab:search_failure}, a large proportion of OVM-guided search failures occur during the selection stage, highlighting it as a critical issue~\footnote{We present only the results of OVM-guided search, as generation failures in PRM-guided search are expected to be similar due to the independence of the generation and selection stages.}.

\begin{table}[h]
\caption{\label{tab:search_failure}Fraction of OVM-guided search failure sources across benchmarks and models (`dsm': `deepseekmath', `mst': `mistral'; `G': `Generation', `S': `Selection').}
\vskip 0.1in
\small
\begin{center}
\setlength{\tabcolsep}{1.5mm}
\begin{tabular}{lrrrrrr}
\toprule
                       & \multicolumn{2}{c}{GSM8K}  & \multicolumn{2}{c}{MATH}         & \multirow{2}{*}{OOD-L4} & \multirow{2}{*}{OOD-L5} \\
                    & dsm & mst     & dsm     & mst       &         &    \\
\midrule
G &                  11.4\% & 16.5\%  & 20.0\%  & 22.9\%    & 15.7\%  & 18.8\%            \\
S &                  88.6\% & 83.5\%  & 80.0\%  & 77.1\%    & 84.3\%  & 81.2\%             \\
\bottomrule
\end{tabular}
\end{center}
\vskip -0.1in
\end{table}

Selection failures in verifier-guided search are directly attributable to verifiers. When verifiers fail to differentiate between valid and invalid paths, and mistakenly assign low ranks to all valid paths, none of them will be further explored, resulting in a selection failure. We refer to this issue as ``verifier failures''. Such failures, which prune all valid paths as failing to select any, ultimately lead to search failures.

To validate the role of verifier failures in contributing to search scaling flaws, we examine the relationship between the success of the selection stage and the number of candidates. Specifically, we analyze the performance of verifier selection in correctly identifying and selecting at least one valid path as the number of candidates increases during the first selection stage. To ensure that the analysis accounts for the presence of valid paths in the candidate set, we use oracle selection performance as a baseline. This baseline serves as a reference for the maximum potential success of the selection process, independent of verifier performance.

\begin{figure*}[!ht]
    \centering
    \subfigbottomskip=2pt
    \subfigcapskip=-2pt
    \subfigure[OVM, MATH (DeepSeekMath 7B)]{
        \includegraphics[width=0.4\linewidth]{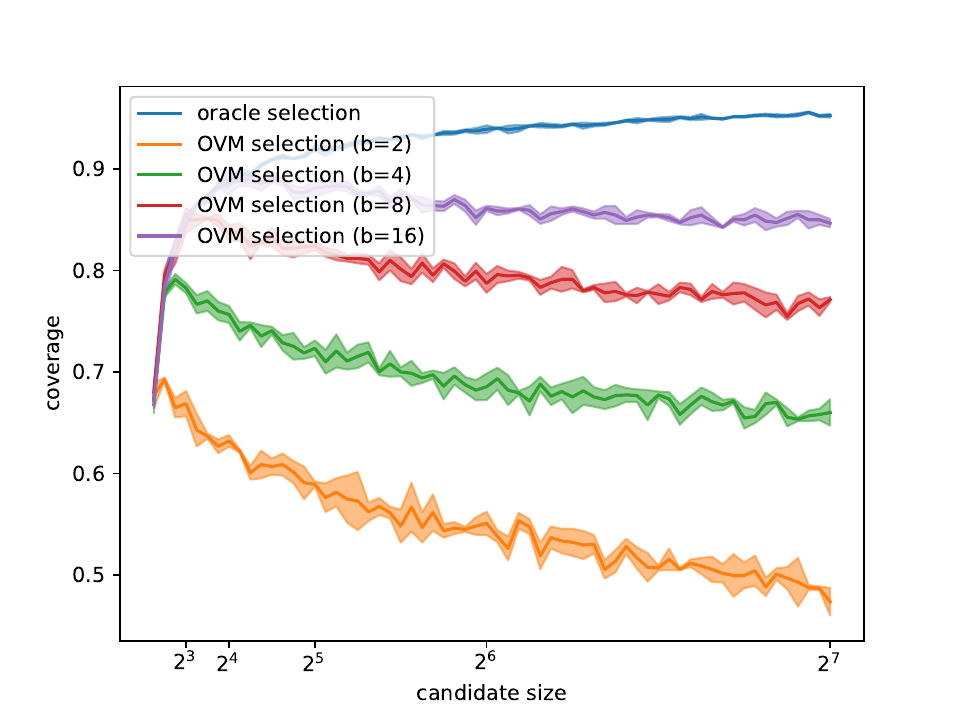}}
    \subfigure[OVM, OOD-L5]{
        \includegraphics[width=0.4\linewidth]{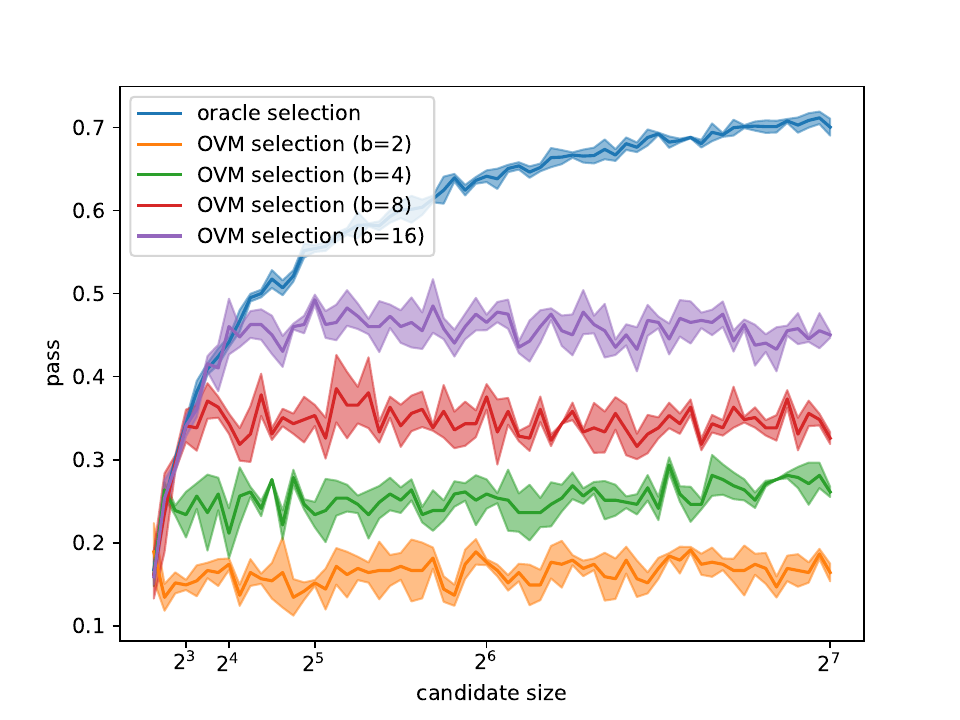}}

    \subfigure[PRM, MATH (DeepSeekMath 7B)]{
        \includegraphics[width=0.4\linewidth]{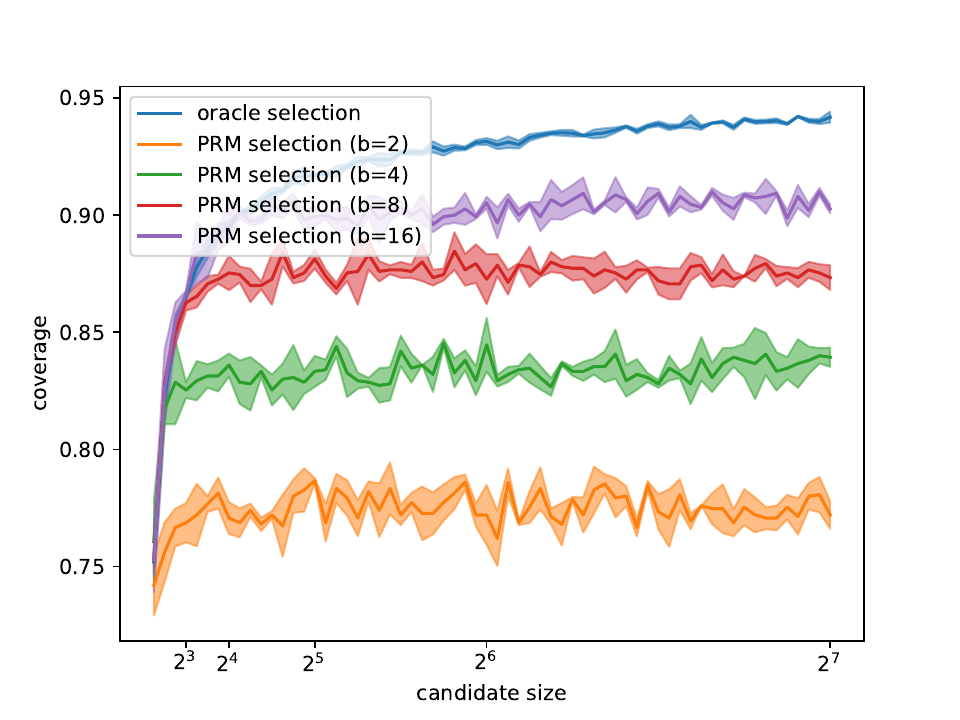}}
    \subfigure[PRM, OOD-L5]{
        \includegraphics[width=0.4\linewidth]{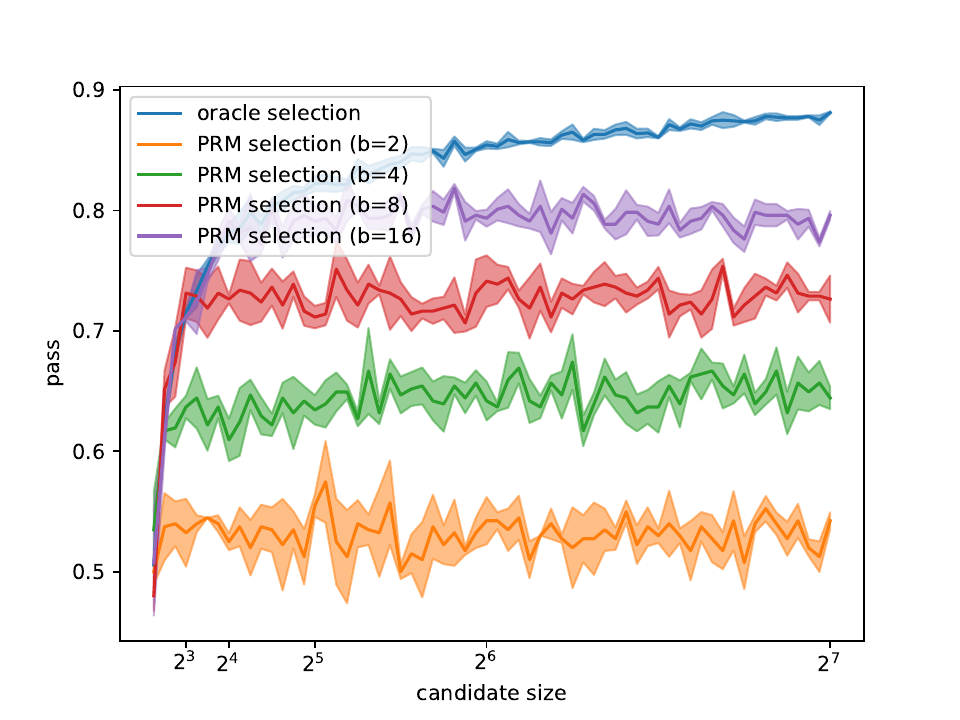}}
    \caption{\label{fig:verifier_flaws}Scaling failures of verifier selection at the first selection stage across various beam sizes on MATH and OOD-L5.}
\end{figure*}

\paragraph{Verifier selection scaling failures} 
\textit{There are verifier selection scaling failures during the selection stage}. As shown in~\cref{fig:verifier_flaws}, verifier selection exhibits scaling failures. Specifically, the performance of verifier selection improves only marginally, saturates, or even decreases as the candidate size increases, despite the presence of valid paths across more problems, as indicated by the oracle selection performance. This phenomenon is consistent across various beam sizes. While selecting and exploring more candidates improves robustness to verifier limitations—evidenced by the reduced gap between verifier selection and oracle selection performance—a significant gap persists even at the largest beam size tested, $b=16$. These scaling failures suggest that verifier selection is a key bottleneck in the success of the selection process, and increasing the candidate size offers limited improvement in addressing this issue.

The scaling failure of verifier selection can explain the diminishing advantage of verifier-guided search. Initially, verifier-guided search is more efficient than repeated sampling, as it effectively selects valid paths and reallocates computational resources for several problems. However, as scaling increases, even though valid paths are available across a broader range of problems, verifiers fail to identify and select them. In contrast, repeated sampling explores more paths without being constrained by verifier failures, ultimately outperforming verifier-guided search at larger scales.

\subsection{More Challenging Scenarios}
\label{sec:challenging_block_more}
In this section, we analyze the failed selection stages during the search, showing that the search process is most hindered when valid paths are sparse.

\paragraph{Sparser candidate space} 
\textit{Verifier selection failures occur and block search more often when valid paths are sparse}. We investigate the failed selection stages during the search and examine the valid path sparsity of these stages. Valid path sparsity is defined as the fraction of valid paths among the candidates. First, we group the valid path sparsity across all selection stages of unsolved problems into four uniform categories. Next, we identify the specific failure stage in each search process where verifier failures occur. we use these groupings to plot the distribution of valid path sparsity across the identified failure stages.

\begin{figure*}[!ht]
    \centering
    \subfigbottomskip=2pt
    \subfigcapskip=-2pt
    \subfigure[MATH]{
        \includegraphics[width=0.4\linewidth]{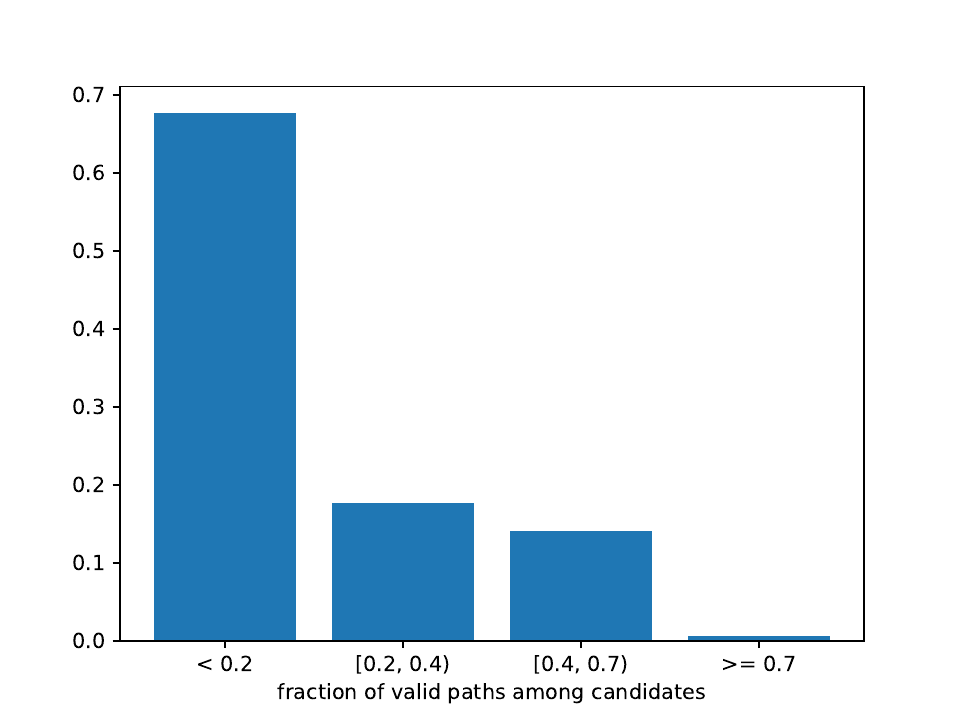}}
    \subfigure[OOD-L5]{
        \includegraphics[width=0.4\linewidth]{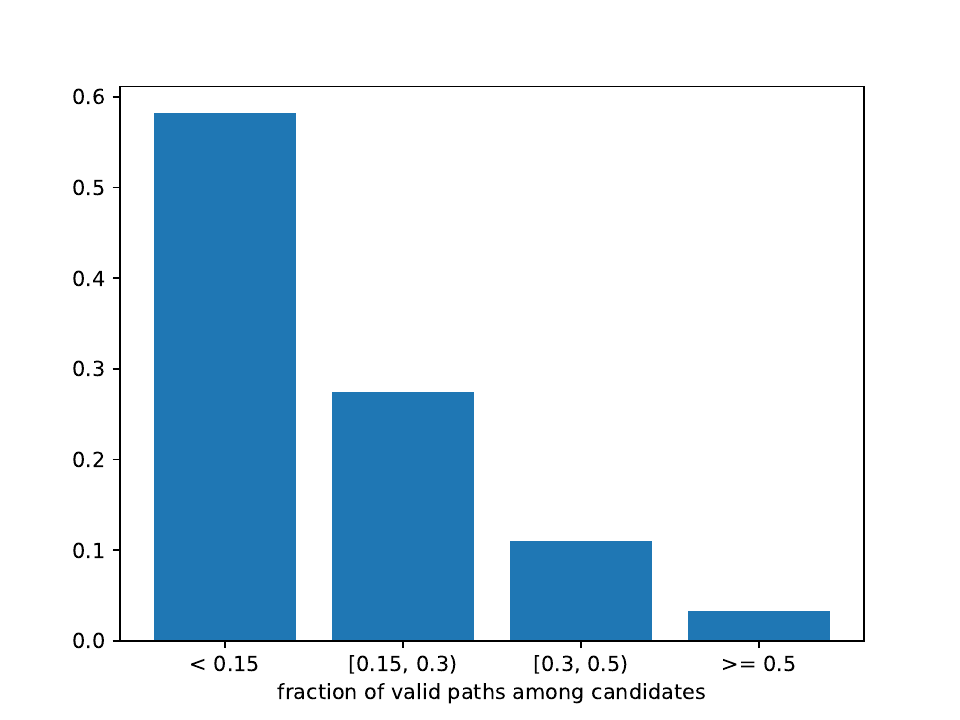}}
    \caption{\label{fig:valid_cand_proportion}Distribution of OVM failures across groups based on valid path sparisty on MATH and OOD-L5 (DeepSeekMath 7B).}
\end{figure*}

As illustrated in~\cref{fig:valid_cand_proportion}, the distribution of failed selection stages demonstrates a monotonic trend: as valid path sparsity decreases, the proportion of failed selection stages increases. This observation aligns with intuition, as identifying valid paths becomes increasingly difficult in sparser candidate spaces.

These findings reveal that verifier failures become increasingly significant during the search process, amplifying the risk of search failure when solving sparser correct solution spaces, where the identification and selection of valid paths become considerably harder.


Although search is expected to offer greater efficiency than repeated sampling in solving more challenging problems by reallocating computational resources through effective selection, our observations suggest that these challenging scenarios are more susceptible to verifier failures, thereby exacerbating scaling flaws. 

\section{Alleviating Verifier Failures}
Imperfect verifiers can lead to verifier failures, obstructing the success of the search process. In this section, we explore two simple methods to alleviate verifier failures by reducing dependency on verifiers: stochastic selection and integration with one-time Monte Carlo rollout.


\paragraph{Experimental setup} 
We evaluate these methods across all the selection stages of the search with $b=8,k=64$. For each method, we measure the accuracy improvement in the selection stage before and after its application.

\paragraph{Stochastic selection} 
Imperfect verifiers can produce incorrect candidate rankings, potentially leading to misguided selection decisions. To mitigate the risk of over-reliance on erroneous rankings, we introduce stochasticity into the selection stage. Rather than deterministically selecting candidates based solely on verifier-predicted score rankings, we apply a softmax function to the candidates' scores and sample from the resulting probability distribution. This approach maintains a preference for high-scoring candidates while still allowing lower-scoring ones a chance to be selected, thereby reducing the risk of incorrectly pruning misranked valid paths.

We experiment with three temperature settings: 0.1, 1, and 10. Higher temperatures reduce reliance on verifier evaluations, being closer to uniform selection. Conversely, lower temperatures increase dependence on verifier evaluations, approximating deterministic selection.

As shown in~\cref{tab:inference_modify_ovm}, stochastic selection improves selection stage accuracy across all benchmarks and models, regardless of temperature, with a notable improvement of up to 11.2\% on OOD-L4 and OOD-L5. Interestingly, on GSM8K, a lower temperature (0.1) yields equal or even greater accuracy gains compared to a higher temperature (10), whereas this trend reverses on MATH and OOD settings. This observation aligns with intuition: since MATH and OOD settings experience more severe verifier failures than GSM8K, reducing reliance on verifier selection through a higher temperature could be more beneficial in these scenarios.


\paragraph{One-time Monte Carlo rollout} 
This method aims to enhance candidate evaluation by incorporating simulated rewards alongside verifier-predicted scores. Specifically, we perform a one-time rollout for each partial path $S^{(1:t)}$ until completion and obtain the reward of the resulting full path.~\footnote{The reward is estimated by the same verifier based on the complete path.} We then linearly combine this reward $r$ with the verifier-predicted score $v^{(1:t)}$ using the formula $\lambda r + (1 - \lambda) v^{(1:t)}$, where $\lambda$ controls the balance between the simulated reward and the verifier’s evaluation.

As shown in~\cref{tab:inference_modify_ovm}, increasing $\lambda$ generally results in higher accuracy gains. Notably, the highest accuracy gain is achieved when relying entirely on the simulated reward, without incorporating the verifier-predicted score. This underscores the limitations of verifiers in candidate evaluation.

\begin{table}[h]
\caption{\label{tab:inference_modify_ovm}Accuracy gain over OVM on the selection stage through two inference-time modification methods (DeepSeekMath 7B).}
\vskip 0.1in
\small
\begin{center}
\begin{tabular}{lrrrr}
\toprule
       & GSM8K       & MATH     & OOD-L4 & OOD-L5 \\
\midrule
\multicolumn{5}{l}{\textit{temperature} in {stochastic selection}} \\
 0.1 & 1.6\%        & 2.8\%         &  8.7\%           & 5.3\%            \\
 1   & 2.0\%        & 5.4\%         &  11.8\%           & 10.3\%            \\
 10   & 1.6\%       & 6.1\%         &  11.2\%           & 11.2\%            \\
\midrule
\multicolumn{5}{l}{\textit{lambda} for {one-time Monte Carlo rollout}} \\
 0.5 & 1.3\%        & -0.8\%         & 2.1\%            & 1.8\%            \\
 0.75 & 1.4\%        & -0.3\%         & 3.0\%            & 2.2\%            \\
 1 & 1.8\%       & 1.7\%         & 6.0\%            &  2.5\%           \\
\bottomrule
\end{tabular}
\end{center}
\vskip -0.1in
\end{table}

\section{Discussion}
\label{sec:discussion}

This work focuses on scaling flaws related to coverage, rather than precision~\cite{RS24}. While precision is important for single-response applications, it is often limited by reward models or selection rules for the final selection. Coverage, however, represents the upper bound of precision and directly equates to it in applications with oracle solution selection, such as automatic theorem proving~\cite{MINIF2F22} and code generation~\cite{codex21}.

\paragraph{Limitations} We do not investigate the impact of scaling verifier sizes and the size of the training dataset. Larger verifier models and more extensive training data could potentially reduce verifier failures and alleviate scaling flaws. 

\paragraph{Future work} A promising direction is to reduce reliance on verifier selection, as discussed in this work. Another avenue is detecting verifier failures and adapting verifier usage accordingly. Uncertainty measures could be useful for identifying these failures.

\section{Conclusion}

We investigate the scaling flaws of verifier-guided search, identifying verifier failures as their underlying cause. While designed to enhance performance on challenging problems, these methods struggle with scalability as problem complexity grows and in real-world OOD settings. Relaxing the reliance on verifier scores could be a promising direction.

\section*{Impact Statements}
This paper presents work whose goal is to advance the field of Machine Learning. There are many potential societal consequences of our work, none which we feel must be specifically highlighted here.

\newpage
\bibliography{custom}

\begin{thebibliography}{20}
\providecommand{\natexlab}[1]{#1}
\providecommand{\url}[1]{\texttt{#1}}
\expandafter\ifx\csname urlstyle\endcsname\relax
  \providecommand{\doi}[1]{doi: #1}\else
  \providecommand{\doi}{doi: \begingroup \urlstyle{rm}\Url}\fi

\bibitem[Brown et~al.(2024)Brown, Juravsky, Ehrlich, Clark, Le, R{\'{e}}, and Mirhoseini]{RS24}
Brown, B. C.~A., Juravsky, J., Ehrlich, R.~S., Clark, R., Le, Q.~V., R{\'{e}}, C., and Mirhoseini, A.
\newblock Large language monkeys: Scaling inference compute with repeated sampling.
\newblock \emph{CoRR}, abs/2407.21787, 2024.
\newblock \doi{10.48550/ARXIV.2407.21787}.
\newblock URL \url{https://doi.org/10.48550/arXiv.2407.21787}.

\bibitem[Chen et~al.(2024)Chen, Liao, Li, and Fan]{AlphaMath24}
Chen, G., Liao, M., Li, C., and Fan, K.
\newblock Alphamath almost zero: process supervision without process.
\newblock \emph{CoRR}, abs/2405.03553, 2024.
\newblock \doi{10.48550/ARXIV.2405.03553}.
\newblock URL \url{https://doi.org/10.48550/arXiv.2405.03553}.

\bibitem[Chen et~al.(2021)Chen, Tworek, Jun, Yuan, de~Oliveira~Pinto, Kaplan, Edwards, Burda, Joseph, Brockman, Ray, Puri, Krueger, Petrov, Khlaaf, Sastry, Mishkin, Chan, Gray, Ryder, Pavlov, Power, Kaiser, Bavarian, Winter, Tillet, Such, Cummings, Plappert, Chantzis, Barnes, Herbert{-}Voss, Guss, Nichol, Paino, Tezak, Tang, Babuschkin, Balaji, Jain, Saunders, Hesse, Carr, Leike, Achiam, Misra, Morikawa, Radford, Knight, Brundage, Murati, Mayer, Welinder, McGrew, Amodei, McCandlish, Sutskever, and Zaremba]{codex21}
Chen, M., Tworek, J., Jun, H., Yuan, Q., de~Oliveira~Pinto, H.~P., Kaplan, J., Edwards, H., Burda, Y., Joseph, N., Brockman, G., Ray, A., Puri, R., Krueger, G., Petrov, M., Khlaaf, H., Sastry, G., Mishkin, P., Chan, B., Gray, S., Ryder, N., Pavlov, M., Power, A., Kaiser, L., Bavarian, M., Winter, C., Tillet, P., Such, F.~P., Cummings, D., Plappert, M., Chantzis, F., Barnes, E., Herbert{-}Voss, A., Guss, W.~H., Nichol, A., Paino, A., Tezak, N., Tang, J., Babuschkin, I., Balaji, S., Jain, S., Saunders, W., Hesse, C., Carr, A.~N., Leike, J., Achiam, J., Misra, V., Morikawa, E., Radford, A., Knight, M., Brundage, M., Murati, M., Mayer, K., Welinder, P., McGrew, B., Amodei, D., McCandlish, S., Sutskever, I., and Zaremba, W.
\newblock Evaluating large language models trained on code.
\newblock \emph{CoRR}, abs/2107.03374, 2021.
\newblock URL \url{https://arxiv.org/abs/2107.03374}.

\bibitem[Cobbe et~al.(2021)Cobbe, Kosaraju, Bavarian, Chen, Jun, Kaiser, Plappert, Tworek, Hilton, Nakano, Hesse, and Schulman]{GSM8K21}
Cobbe, K., Kosaraju, V., Bavarian, M., Chen, M., Jun, H., Kaiser, L., Plappert, M., Tworek, J., Hilton, J., Nakano, R., Hesse, C., and Schulman, J.
\newblock Training verifiers to solve math word problems.
\newblock \emph{CoRR}, abs/2110.14168, 2021.
\newblock URL \url{https://arxiv.org/abs/2110.14168}.

\bibitem[Hao et~al.(2023)Hao, Gu, Ma, Hong, Wang, Wang, and Hu]{RAP23}
Hao, S., Gu, Y., Ma, H., Hong, J.~J., Wang, Z., Wang, D.~Z., and Hu, Z.
\newblock Reasoning with language model is planning with world model.
\newblock In Bouamor, H., Pino, J., and Bali, K. (eds.), \emph{Proceedings of the 2023 Conference on Empirical Methods in Natural Language Processing, {EMNLP} 2023, Singapore, December 6-10, 2023}, pp.\  8154--8173. Association for Computational Linguistics, 2023.
\newblock \doi{10.18653/V1/2023.EMNLP-MAIN.507}.
\newblock URL \url{https://doi.org/10.18653/v1/2023.emnlp-main.507}.

\bibitem[Hendrycks et~al.(2021)Hendrycks, Burns, Kadavath, Arora, Basart, Tang, Song, and Steinhardt]{MATH21}
Hendrycks, D., Burns, C., Kadavath, S., Arora, A., Basart, S., Tang, E., Song, D., and Steinhardt, J.
\newblock Measuring mathematical problem solving with the {MATH} dataset.
\newblock In Vanschoren, J. and Yeung, S. (eds.), \emph{Proceedings of the Neural Information Processing Systems Track on Datasets and Benchmarks 1, NeurIPS Datasets and Benchmarks 2021, December 2021, virtual}, 2021.
\newblock URL \url{https://datasets-benchmarks-proceedings.neurips.cc/paper/2021/hash/be83ab3ecd0db773eb2dc1b0a17836a1-Abstract-round2.html}.

\bibitem[Jiang et~al.(2023)Jiang, Sablayrolles, Mensch, Bamford, Chaplot, de~Las~Casas, Bressand, Lengyel, Lample, Saulnier, Lavaud, Lachaux, Stock, Scao, Lavril, Wang, Lacroix, and Sayed]{Mistral7B-23}
Jiang, A.~Q., Sablayrolles, A., Mensch, A., Bamford, C., Chaplot, D.~S., de~Las~Casas, D., Bressand, F., Lengyel, G., Lample, G., Saulnier, L., Lavaud, L.~R., Lachaux, M., Stock, P., Scao, T.~L., Lavril, T., Wang, T., Lacroix, T., and Sayed, W.~E.
\newblock Mistral 7b.
\newblock \emph{CoRR}, abs/2310.06825, 2023.
\newblock \doi{10.48550/ARXIV.2310.06825}.
\newblock URL \url{https://doi.org/10.48550/arXiv.2310.06825}.

\bibitem[Kwon et~al.(2023)Kwon, Li, Zhuang, Sheng, Zheng, Yu, Gonzalez, Zhang, and Stoica]{VLLM23}
Kwon, W., Li, Z., Zhuang, S., Sheng, Y., Zheng, L., Yu, C.~H., Gonzalez, J., Zhang, H., and Stoica, I.
\newblock Efficient memory management for large language model serving with pagedattention.
\newblock In Flinn, J., Seltzer, M.~I., Druschel, P., Kaufmann, A., and Mace, J. (eds.), \emph{Proceedings of the 29th Symposium on Operating Systems Principles, {SOSP} 2023, Koblenz, Germany, October 23-26, 2023}, pp.\  611--626. {ACM}, 2023.
\newblock \doi{10.1145/3600006.3613165}.
\newblock URL \url{https://doi.org/10.1145/3600006.3613165}.

\bibitem[Lightman et~al.(2024)Lightman, Kosaraju, Burda, Edwards, Baker, Lee, Leike, Schulman, Sutskever, and Cobbe]{PRM24}
Lightman, H., Kosaraju, V., Burda, Y., Edwards, H., Baker, B., Lee, T., Leike, J., Schulman, J., Sutskever, I., and Cobbe, K.
\newblock Let's verify step by step.
\newblock In \emph{The Twelfth International Conference on Learning Representations, {ICLR} 2024, Vienna, Austria, May 7-11, 2024}. OpenReview.net, 2024.
\newblock URL \url{https://openreview.net/forum?id=v8L0pN6EOi}.

\bibitem[Loshchilov \& Hutter(2019)Loshchilov and Hutter]{AdamW19}
Loshchilov, I. and Hutter, F.
\newblock Decoupled weight decay regularization.
\newblock In \emph{7th International Conference on Learning Representations, {ICLR} 2019, New Orleans, LA, USA, May 6-9, 2019}. OpenReview.net, 2019.
\newblock URL \url{https://openreview.net/forum?id=Bkg6RiCqY7}.

\bibitem[Shao et~al.(2024)Shao, Wang, Zhu, Xu, Song, Zhang, Li, Wu, and Guo]{DeepSeekMath24}
Shao, Z., Wang, P., Zhu, Q., Xu, R., Song, J., Zhang, M., Li, Y.~K., Wu, Y., and Guo, D.
\newblock Deepseekmath: Pushing the limits of mathematical reasoning in open language models.
\newblock \emph{CoRR}, abs/2402.03300, 2024.
\newblock \doi{10.48550/ARXIV.2402.03300}.
\newblock URL \url{https://doi.org/10.48550/arXiv.2402.03300}.

\bibitem[Snell et~al.(2024)Snell, Lee, Xu, and Kumar]{Snell24}
Snell, C., Lee, J., Xu, K., and Kumar, A.
\newblock Scaling {LLM} test-time compute optimally can be more effective than scaling model parameters.
\newblock \emph{CoRR}, abs/2408.03314, 2024.
\newblock \doi{10.48550/ARXIV.2408.03314}.
\newblock URL \url{https://doi.org/10.48550/arXiv.2408.03314}.

\bibitem[Tian et~al.(2024)Tian, Peng, Song, Jin, Yu, Mi, and Yu]{tian24}
Tian, Y., Peng, B., Song, L., Jin, L., Yu, D., Mi, H., and Yu, D.
\newblock Toward self-improvement of llms via imagination, searching, and criticizing.
\newblock \emph{CoRR}, abs/2404.12253, 2024.
\newblock \doi{10.48550/ARXIV.2404.12253}.
\newblock URL \url{https://doi.org/10.48550/arXiv.2404.12253}.

\bibitem[Wan et~al.(2024)Wan, Feng, Wen, McAleer, Wen, Zhang, and Wang]{Wan24}
Wan, Z., Feng, X., Wen, M., McAleer, S.~M., Wen, Y., Zhang, W., and Wang, J.
\newblock Alphazero-like tree-search can guide large language model decoding and training.
\newblock In \emph{Forty-first International Conference on Machine Learning, {ICML} 2024, Vienna, Austria, July 21-27, 2024}. OpenReview.net, 2024.
\newblock URL \url{https://openreview.net/forum?id=C4OpREezgj}.

\bibitem[Wang et~al.(2024)Wang, Li, Shao, Xu, Dai, Li, Chen, Wu, and Sui]{Math-Shepherd24}
Wang, P., Li, L., Shao, Z., Xu, R., Dai, D., Li, Y., Chen, D., Wu, Y., and Sui, Z.
\newblock Math-shepherd: Verify and reinforce llms step-by-step without human annotations.
\newblock In Ku, L., Martins, A., and Srikumar, V. (eds.), \emph{Proceedings of the 62nd Annual Meeting of the Association for Computational Linguistics (Volume 1: Long Papers), {ACL} 2024, Bangkok, Thailand, August 11-16, 2024}, pp.\  9426--9439. Association for Computational Linguistics, 2024.
\newblock \doi{10.18653/V1/2024.ACL-LONG.510}.
\newblock URL \url{https://doi.org/10.18653/v1/2024.acl-long.510}.

\bibitem[Wu et~al.(2024{\natexlab{a}})Wu, Sun, Li, Welleck, and Yang]{rebase24}
Wu, Y., Sun, Z., Li, S., Welleck, S., and Yang, Y.
\newblock An empirical analysis of compute-optimal inference for problem-solving with language models.
\newblock \emph{CoRR}, abs/2408.00724, 2024{\natexlab{a}}.
\newblock \doi{10.48550/ARXIV.2408.00724}.
\newblock URL \url{https://doi.org/10.48550/arXiv.2408.00724}.

\bibitem[Wu et~al.(2024{\natexlab{b}})Wu, Sun, Li, Welleck, and Yang]{wu2024inference}
Wu, Y., Sun, Z., Li, S., Welleck, S., and Yang, Y.
\newblock Inference scaling laws: An empirical analysis of compute-optimal inference for problem-solving with language models.
\newblock \emph{arXiv preprint arXiv:2408.00724}, 2024{\natexlab{b}}.

\bibitem[Xin et~al.(2024)Xin, Ren, Song, Shao, Zhao, Wang, Liu, Zhang, Lu, Du, Gao, Zhu, Yang, Gou, Wu, Luo, and Ruan]{deepseekprover15-24}
Xin, H., Ren, Z.~Z., Song, J., Shao, Z., Zhao, W., Wang, H., Liu, B., Zhang, L., Lu, X., Du, Q., Gao, W., Zhu, Q., Yang, D., Gou, Z., Wu, Z.~F., Luo, F., and Ruan, C.
\newblock Deepseek-prover-v1.5: Harnessing proof assistant feedback for reinforcement learning and monte-carlo tree search.
\newblock \emph{CoRR}, abs/2408.08152, 2024.
\newblock \doi{10.48550/ARXIV.2408.08152}.
\newblock URL \url{https://doi.org/10.48550/arXiv.2408.08152}.

\bibitem[Yu et~al.(2024)Yu, Gao, and Wang]{OVM23}
Yu, F., Gao, A., and Wang, B.
\newblock Ovm, outcome-supervised value models for planning in mathematical reasoning.
\newblock In Duh, K., G{\'{o}}mez{-}Adorno, H., and Bethard, S. (eds.), \emph{Findings of the Association for Computational Linguistics: {NAACL} 2024, Mexico City, Mexico, June 16-21, 2024}, pp.\  858--875. Association for Computational Linguistics, 2024.
\newblock \doi{10.18653/V1/2024.FINDINGS-NAACL.55}.
\newblock URL \url{https://doi.org/10.18653/v1/2024.findings-naacl.55}.

\bibitem[Zheng et~al.(2022)Zheng, Han, and Polu]{MINIF2F22}
Zheng, K., Han, J.~M., and Polu, S.
\newblock minif2f: a cross-system benchmark for formal olympiad-level mathematics.
\newblock In \emph{The Tenth International Conference on Learning Representations, {ICLR} 2022, Virtual Event, April 25-29, 2022}. OpenReview.net, 2022.
\newblock URL \url{https://openreview.net/forum?id=9ZPegFuFTFv}.

\end{thebibliography}
\bibliographystyle{icml2025}

\newpage
\appendix
\onecolumn
\section{Appendix}

\begin{table}[h]
\centering
\caption{Summary of Notations Used in the Paper}
\resizebox{0.8\textwidth}{!}{%
\begin{tabular}{cl}
\toprule
\textbf{Notation} & \textbf{Description} \\ 
\midrule
$q$ & Mathematical reasoning question requiring a sequence of steps \\
$S$ & Solution path for a question, $S=[s^1, \dots, s^T,a]$ \\
$s^i$ & $i$-th step in a solution path \\
$a$ & Final answer in a solution path \\
$T$ & Number of steps in a solution path \\
$y$ & Binary label (0 or 1) indicating the correctness of $a$ \\
$S^{(1:t)}$ & Partial solution path up to step $t$, $S^{(1:t)}=[s^1, \dots, s^t]$ \\
$\mathbb{S}^{(1:t)}$ & Set of candidate partial paths $\mathbb{S}^{(1:t)} = \{S_k^{(1:t)}\}_{k=1}^{K}$ \\
$v^{(1:t)}$ & The score for the partial path $S^{(1:t)}$ \\
$\mathbb{V}^{(1:t)}$ & Set of scores for candidate partial paths $\mathbb{V}^{(1:t)} = \bigl\{v^{(1:t)}_k \bigl\}_{k=1}^{K}$ \\
$K$ & Number of candidates \\
$b$ & Beam size \\
\bottomrule
\end{tabular}%
}
\label{tab:notations}
\end{table}

\subsection{Verifier Training}
\label{app:vm_training}

\paragraph{OVM training dataset construction} 
OVMs are trained on automatically constructed datasets, where the correctness of the final answer serves as the label for each instance. The training dataset is constructed from the generator and the given question-answer pairs: For each pair $(q,a)\in\mathcal{Q}$, the generator produces $n$ solution paths, resulting in $|\mathcal{Q}|\times n$ question-solution pairs. The label $y$ for each solution $S$ is determined by checking the correctness of the final answer, e.g. matching it to the ground truth $a$, with 1 indicating ``correct'' and 0 indicating ``incorrect''. This process generates a training dataset of $(q, S, y)$ tuples for value models.

\paragraph{PRM training dataset} 
PRMs are trained at a fine-grained step level, requiring annotations of step correctness. In this study, we use the open-source Math-Shepherd process data~\cite{Math-Shepherd24} to train the PRMs. 

Both OVMs and PRMs are trained with mean squared losses.

\subsection{Step-Level Beam Search}

The algorithm is shown in~\cref{algo:beam_search}.

\begin{algorithm}[h]
\small
\caption{\label{algo:beam_search}Step-Level Beam Search}
\begin{algorithmic}[1]

\Statex $\textbf{Input:}$ Question $q$, Beam size $b$, Sampled steps per state $K$, Maximum step count $T^{max}$
\Statex $\textbf{Output:}$ $b$ solution sequences for $q$
\Statex $\textbf{Model:}$ $\operatorname{Generator}$ and $\operatorname{VM}$

    \State Initialize step sequences $\mathbb{S} \gets \{\}$
    \State Sample initial steps $\{s_1^1,\dots,s_K^1\}$
    \State Select $b$ steps via \Call{Selection}{$q$, $\{s_1^1,\dots,s_K^1\}$, $b$, $\operatorname{VM}$} and add to $\mathbb{S}$
    \State $t \gets 1$
    \While{sequences in $\mathbb{S}$ are not complete and $t < T^{max}$}
        \State $\mathbb{S}_{\text{new}} \gets \{\}$
        \For{each sequence $S^{(1:t)}$ in $\mathbb{S}$}
            \For{$i = 1$ to $K/b$}
                \State $S^{(1:t+1)}_i=\operatorname{Generator}(S_i^{(1:t)};q)$
                \State $\mathbb{S}_{\text{new}} \gets \mathbb{S}_{\text{new}}+S^{(1:{t+1})}_i$
            \EndFor
        \EndFor

        \State $\mathbb{S}_{\text{new}} \gets$ \Call{Selection}{$q$, $\mathbb{S}_{\text{new}}$, $b$, $\operatorname{VM}$}
        \State $\mathbb{S} \gets \mathbb{S}_{\text{new}}$
        \State $t \gets t+1$
    \EndWhile
\Statex \Return $\mathbb{S}$
\end{algorithmic}

\end{algorithm}

\subsection{Implementation Details}
\label{app:implementation}

\subsubsection{OVMs}
\label{app:ovm_implementation}
\paragraph{Training generators} 
We train the base models (Mistral 7B or DeepSeekMath 7B) on the training sets of each setting. In MATH, we split the steps using period and newline characters. We normalize datasets to use the newline character as the marker for the end of each step across all tasks. In all datasets, supervised fine-tuning is performed for 2 epochs with a batch size of 128. We use a linear learning rate scheduler with a maximum learning rate of 2e-6 for Mistral 7B and 5e-5 for DeepSeekMath 7B. The AdamW optimizer~\cite{AdamW19} is used for training. 

\paragraph{Building training dataset for OVMs} 
The dataset construction process is introduced in ~\cref{app:vm_training}. We sample 50 solution paths per problem in GSM8K, and 100 solution paths per problem in MATH. For GSM8K, we follow the setup in~\cite{OVM23}, with a decoding temperature of 0.7 and top-k set to 50 for dataset collection. The maximum new token length is set to 400 for GSM8K. In MATH (including OOD settings), we use a decoding temperature of 1, top-p of 0.98, and a maximum new token length of 2000. As token sequences in MATH are long, we apply vllm~\cite{VLLM23} to accelerate the generation process.

\paragraph{Training OVMs} 
OVMs are initialized from the corresponding generator checkpoints and trained for one epoch, using the same learning rate scheduler as the generator training. The batch size is set to 128 in GSM8K and to 512 in MATH. The optimizer used for training is AdamW.

\subsubsection{PRMs}
We use the open-source Math-Shepherd dataset~\cite{Math-Shepherd24} to train both the generators and PRMs. 

\paragraph{Data extraction} 
We extract training problems for each setting. Specifically, for the GSM8K task, we extract all problems from the training split of GSM8K, and for the MATH task, we extract all problems from the training split of MATH. Similar extractions are performed for the OOD settings.

\paragraph{Data preprocessing} 
Since the data format in Math-Shepherd is inconsistent, we normalize the solution paths. We detect steps in each path, normalize them to be split by a newline character, and summarize the final answer in the format of ``The answer is xx''. For MATH problems, the final answer is enclosed in ``$\backslash$boxed\{\}''.

\paragraph{Training generators} 
For each setting, we randomly select one correct solution for each training problem. If no correct solution is provided, we randomly select one other solution. The training parameters, including the number of epochs, learning schedule, batch size, and optimizer for each base model (Mistral 7B or DeepSeekMath 7B), are the same as those in~\cref{app:ovm_implementation}.

\paragraph{Training PRMs} 
We use all solution paths and annotations provided in Math-Shepherd to train PRMs, which are initialized from the corresponding generator checkpoints and trained for one epoch. Same as above, the batch size is set to 128 in GSM8K and 512 in MATH, and AdamW is used for training. The maximum learning rates for Mistral 7B and DeepSeekMath 7B are 2e-6 and 5e-5, respectively.

We observe that the Math-Shepherd data is noisy and some steps are missing labels. We speculate that this might contribute to its inferior performance compared to OVM in this work.

\subsubsection{Step-level beam search} 
In GSM8K, we set the decoding temperature to 0.7, top-k to 50, maximum new token length to 400, and maximum number of steps to 10. In MATH, we set the decoding temperature to 1.0, top-p to 0.98, maximum new token length to 2000, and maximum number of steps to 30. During the beam search process, we prioritize selecting non-duplicate steps. We use vllm in MATH to accelerate token sequence generation.



\end{document}